\documentclass[journal]{vgtc}        
\ifpdf%                                % if we use pdflatex
  \pdfoutput=1\relax                   % create PDFs from pdfLaTeX
  \usepackage{graphicx}                % allow us to embed graphics files
  \DeclareGraphicsExtensions{.pdf,.png,.jpg,.jpeg} % for pdflatex we expect .pdf, .png, or .jpg files
\else%                                 % else we use pure latex
  \usepackage{graphicx}                % allow us to embed graphics files
  \DeclareGraphicsExtensions{.eps}     % for pure latex we expect eps files
\fi%

\graphicspath{{figures/}{pictures/}{images/}{./}} % where to search for the images

\usepackage{microtype}                 % use micro-typography (slightly more compact, better to read)
\PassOptionsToPackage{warn}{textcomp}  % to address font issues with \textrightarrow
\usepackage{textcomp}                  % use better special symbols
\usepackage{mathptmx}                  % use matching math font
\usepackage{times}                     % we use Times as the main font
\usepackage{cite}                      % needed to automatically sort the references
\usepackage{tabu}                      % only used for the table example
\usepackage{booktabs}                  % only used for the table example
\usepackage{amsmath}
\usepackage{amssymb}
\usepackage{mathrsfs}

\usepackage{multirow} 
\usepackage{multicol} 
\usepackage{arydshln}

\usepackage{color}
\newcommand{\lingbo}[1]{}

\newcommand{\etal}{\textit{et al}.}
\onlineid{1670}

\vgtccategory{Research}
\vgtcinsertpkg
\title{Learning Variational Motion Prior for Video-based Motion Capture}
\vspace{10pt}

\author{Xin Chen$^{1*}$, Zhuo Su$^{1*}$, Lingbo Yang$^{1*}$, Pei Cheng$^{1}$, Lan Xu$^{2}$, Bin Fu$^{1}$, and Gang Yu$^{1\dagger}$}
\authorfooter{
\item
 $^{*}$ Joint first authors.
\item
 $^{\dagger}$ corresponding authors.
\item
 $^{1}$ Tencent PCG, Shanghai, China
\item
 $^{2}$ School of Information Science and Technology, ShanghaiTech University, Shanghai, China.
\item
 Xin Chen and Lan Xu, \{chenxin2, xulan1\}@shanghaitech.edu.cn; Zhuo Su, suzhuo13@gmail.com; Lingbo Yang, Pei Cheng and Bin Fu, \{lingboyang, peicheng, brianfu\}@tencent.com; Gang Yu, skicy@outlook.com.
}

\shortauthortitle{Biv \MakeLowercase{\textit{et al.}}: Global Illumination for Fun and Profit}
\abstract{

Motion capture from a monocular video is fundamental and crucial for us humans to naturally experience and interact with each other in Virtual Reality (VR) and Augmented Reality (AR). However, existing methods still struggle with challenging cases involving self-occlusion and complex poses due to the lack of effective motion prior modeling. 
In this paper, we present a novel variational motion prior (VMP) learning approach for video-based motion capture to resolve the above issue. Instead of directly building the correspondence between the video and motion domain, We propose to learn a generic latent space for capturing the prior distribution of all natural motions, which serve as the basis for subsequent video-based motion capture tasks. To improve the generalization capacity of prior space, we propose a transformer-based variational autoencoder pretrained over marker-based 3D mocap data, with a novel style-mapping block to boost the generation quality. Afterward, a separate video encoder is attached to the pretrained motion generator for end-to-end fine-tuning over task-specific video datasets. Compared to existing motion prior models, our VMP model serves as a motion rectifier that can effectively reduce temporal jittering and failure modes in frame-wise pose estimation, leading to temporally stable and visually realistic motion capture results. Furthermore, our VMP-based framework models motion at sequence level and can directly generate motion clips in the forward pass, achieving real-time motion capture during inference. Extensive experiments over both public datasets and in-the-wild videos have demonstrated the efficacy and generalization capability of our framework.
} % end of abstract

\CCScatlist{ % not used in journal version
 \CCScat{I.3.6}{Computing Methodologies}%
 {COMPUTER GRAPHICS}{Methodology and Techniques}
 \CCScat{I.2.6}{Computing Methodologies}%
 {ARTIFICIAL INTELLIGENCE}{Learning}
}

\teaser{
  \centering
  \vspace{10pt}
  \includegraphics[width=\linewidth]{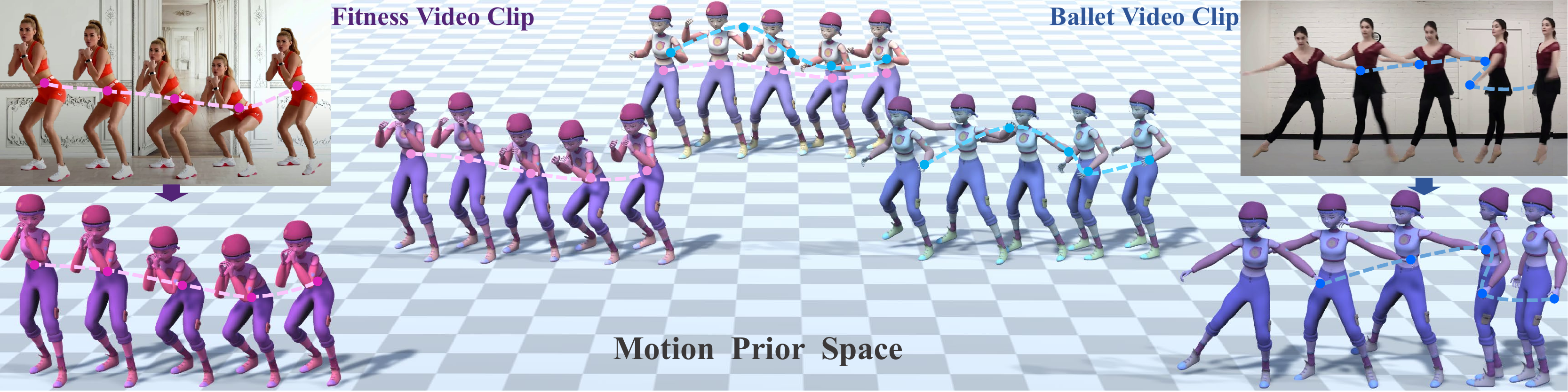}
  \caption{Our transformer-based learning framework achieves robust 3D motion capture from two input video clips (fitness and ballet) by utilizing a novel variational motion prior. This motion prior helps to achieve plausible motion reconstruction from our expressive latent motion prior space. }
  \label{fig:teaser}
}

\begin{document}

%% The ``\maketitle'' command must be the first command after the
%% ``\begin{document}'' command. It prepares and prints the title block.

%% the only exception to this rule is the \firstsection command
\firstsection{Introduction}
\label{sec:intro}

\maketitle
% 1. Problem definition, why we need capture and style synthesis together?
Human motion capture is a fundamental problem in virtual and augmented reality (VR/AR) that benefits various media applications, such as immersive telepresence, game character animation, and meta-universe. 
% 手K = keyframing
% stylized 或许在这里不大合适
Traditional marker-based motion capture methods can produce highly accurate %\cx{and stylized} 
human motion, yet are usually labor-intensive and require sophisticated equipment. 
% Such applications usually involves of motion capture and post-editing, where motion sequences are first captured by sampling movements of human actors and then edited manually in animation software for refinement and style control. 
With the development of deep neural networks, markerless motion capture frameworks that directly reconstruct 3D human motion from a monocular 2D video have attracted increasing attention~\cite{Kocabas_2020_CVPR,Kanazawa_2019_CVPR,wan2021encoder,chen2021sportscap}. As an inherently ill-posed 2D-to-3D estimation problem, existing works typically struggle with complex scenes involving heavy self-occlusion or uncommon activities. How to further incorporate human motion prior knowledge into the monocular capture framework remains challenging.
% Despite the efforts over the past decades on human motion modeling and conditional synthesis~\cite{guo2020action2motion,li2022ganimator}, how to achieve both accurate reconstruction and flexible editing within a single framework remains a considerable challenge.

% 2. Existing works, highlighting prior formulation
% 关键，从conditional motion prior的角度重新叙述motion synthesis的工作，
Motion prior modeling is a long-standing problem in motion capture. Early works attempt to capture the statistical prior using principle components~\cite{Ormoneit2000LearningAT}, autoregressive models~\cite{Taylor2006ModelingHM}, Gaussian mixture models SMPLify~\cite{keepitSMPL} or variational human pose prior~\cite{SMPL-X:2019}, as well as physical constraints such as foot contact~\cite{Brubaker2009EstimatingCD, rempe2020contact} and kinematics~\cite{Li2019Estimating3M, PhysCapTOG2020, Li2021HybrIKAH} to derive regularization terms. Yet hand-crafted prior is usually task-specific with limited applicability.
More recent works typically learn the prior distribution of human motion in a data-driven fashion using deep generative models.
% VIBE & AMP, cited
VIBE~\cite{Kocabas_2020_CVPR} introduces a generative adversarial network with a motion discriminator to distinguish between real 3D MoCap data and synthesized ones, and a similar idea is also adopted for physics-based character control~\cite{Peng2021AMP}.
%Class-conditional motion prior model~\cite{guo2020action2motion, petrovich2021action} utilizes a class type label (e.g. action types) to model the conditional prior of motion under specific action category, . Yet such class-conditional prior setup is often hard to cope with unseen categories, such as sport activities
Another line of methods adopt a variational autoencoder (VAE) to learn an explicit motion representation. VPoser~\cite{SMPL-X:2019} utilize VAE to model human poses without temporal information, and Rempe \etal~\cite{rempe2021humor} further model the distribution of pose changes conditioned on the current pose, achieving robust 3D pose estimation results. However, the pose-level prior model cannot fully capture the holistic temporal motion characteristics. ACTOR~\cite{petrovich2021action} introduces a temporal VAE to learn motion prior at the sequence level, but the prior model is conditioned on discrete action labels such as walking or running, with limited generalization capabilities to unseen action categories~\cite{chen2021sportscap}.

To tackle the above issues, we present a novel variational motion prior (VMP) model that directly captures the holistic characteristics of human motion at the sequence-level. As shown in Fig.~\ref{fig:teaser}, each motion sequence is represented as a single point on the latent prior space, allowing novel motion clips to be generated directly from latent codes with a single forward pass. 
Furthermore, the continuous latent space structure of VMP can support better generalization over unseen motions, for example, the intermediate motion between fitness exercise and ballet dance.
% Unlike existing prior models that characterize motion at frame-level~\cite{guo2020action2motion, rempe2021humor}, our latent space.
To incorporate motion prior in video-based motion capture, we design a two-stage training strategy to build and utilize motion prior respectively. 
During stage I, we propose a transformer-based variational autoencoder to learn the VMP latent space, with a novel non-linear mapping block to boost the expressive power of latent codes via adaptive instance normalization (AdaIN)~\cite{Adain}. 
The prior model is trained in an over a large-scale 3D motion capture database~\cite{AMASS:ICCV:2019} without the supervsion of motion-video pairs.
During stage II, a video encoder with spatial-temporal transformer architecture is introduced to encode the input video clip into the pretrained VMP latent space by extracting and aggregating multi-scale feature pyramids with a high-to-low resolution CNN backbone. 
We argue that the pretrained VMP is crucial for accurate motion reconstruction as the pretrained VMP serves as an effective rectifier that reduce temporal jittering caused by unstable per-frame feature extraction. 
Empirically, VMP brings over 10mm error reduction over 3DPW dataset (in terms of PA-MPJPE, see Table \ref{table:eval}), setting up new performance records in video-based motion capture. Furthermore, VMP also improves the generalization capability of the motion capture framework over in-the-wild videos.

% 5. Our technical contributions: framework; motion vae?; motion prior
In summary, our contributions are threefold:
\begin{itemize} 
% \vspace{-10pt}
% \setlength\itemsep{0em}
    % pipeline: end-to-end transformer-based learning framework for both motion capture and motion sysnthesis
    % to simultaneously xx motion capture and motion sysnthesis
    %
    % 这句不够突出end-to-end learning framework
    % 同时解决两个关联任务的描述 不够流畅 --> a unified framework
    % first 要谦虚吗
    % 0515 ylb. 谦～虚～
    \item We propose a novel variational motion prior (VMP) learning approach for video-based motion capture. Specifically, VMP is implemented as a transformer-based variational autoencoder pretrained over large-scale 3D motion data, providing an expressive latent space for human motion at sequence level.
    \item We propose several effective technical contributions including a novel motion generator, a two-stage prior training strategy and loss objective design to boost the accuracy of motion capture.
    \item We conduct extensive experiments over both academic datasets and in-the-wild videos to validate the efficacy and versatility of the proposed framework, and perform a through ablation study to verify our technical contributions.
    
    % VMP: variational motion prior
    % sz: 加强和第2点的区分度？
    % cx: 目前只有这个思路 问问俞刚老师和凌波意见
    % \item  We introduce a , which is carefully trained from image-based motion capture framework and marker-based motion VAE framework, providing 
    % cx: 同样 是否形容角度重复？
    % natural and plausible motion generation.
    
\end{itemize}

\begin{figure*}[t]
    \centering
    \includegraphics[width=\linewidth]{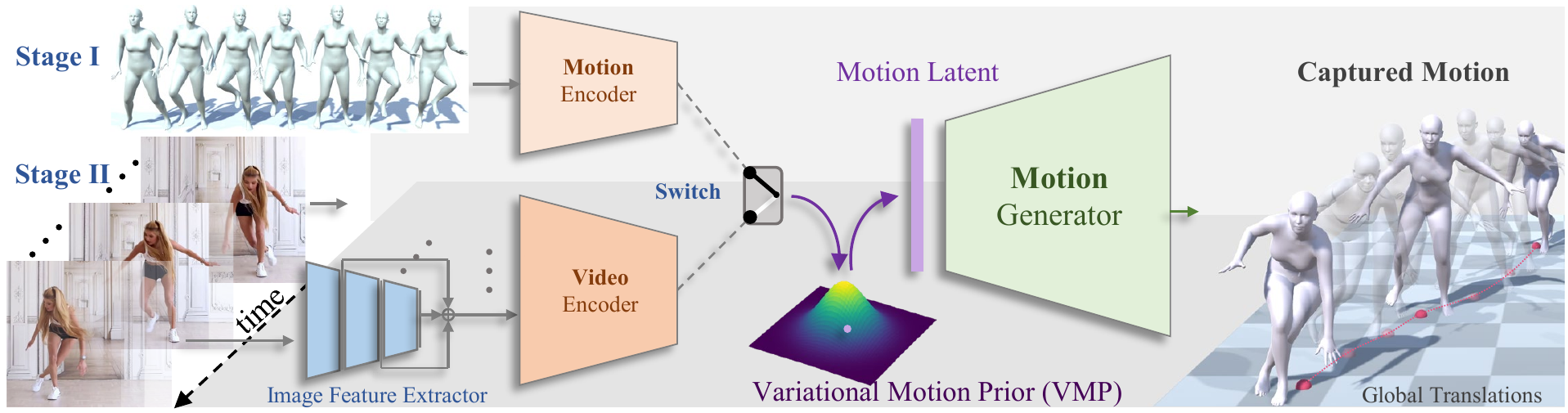}
    \caption{\lingbo{Explicitly draw both motion encoder and video encoder?} Overview of our framework for the video-based motion capture. It consists of a novel-designed variational motion prior (Sec.\ref{sec:vae}) and a transformer-based video encoder (Sec.\ref{sec:cap}),
    % modify to new 3.1 title here @ylb
    which leverages this motion prior to generate natural motions conditioned on an input video. For the supervision, we use a two-stage training strategy with different inputs from motion-only or video-motion dataset. 
    % Here, ${1024,512,256}$ represents the high-to-low image features.
    }
    \label{fig:pipeline_all}
\end{figure*}

% human motion survey
% refer to the survey
% https://doc.weixin.qq.com/doc/w3_AVkA7wbdAN4TLKPSigqTm6yzPCSFB?scode=AJEAIQdfAAoeZkcEhS

\section{Related Work}
\label{sec:related_work}

% to do: add more related work

In this section, we briefly review the representative works in related domain, including human motion capture and motion prior models.
We first provide a brief summary of recent progress on the motion capture technologies, followed by an overview of human motion models.

\hspace*{\fill} \\{\bf Human Motion Capture.}
% marker-based --> early marker-less
Industrial solutions for human motion capture are always marker-based such as~\cite{VICON,Xsens,Vlasic2007}, while sophisticated marker-less capture systems  ~\cite{AguiaSTAST2008,TheobASST2010, UnstructureLan, dou-siggraph2016} turn to use multi-view cameras to alleviate the need for body-worn markers.
Recently, human motion capture technoligies using single-view input have made much progress. 
% skeleton pose
With the aid of deep neural network, single-view methods~\cite{OpenPose,Mehta2017} can estimate 2D/3D human skeletons.
% parametric model
Many general human parametric models~\cite{SCAPE2005,SMPL2015,SMPL-X:2019,STAR_ECCV2020} learned from 3D scans factorize human deformation into pose and shape components. 
% single rgb image
Using these parametric models, motion capture methods can also optionally estimate the body shape.
Pose capture methods~\cite{keepitSMPL, HMR18,joo2020eft, Lassner2017CVPR,chen2021sportscap, kolotouros2019learning,SMPL-X:2019, Kocabas_2021_ICCV} estimation SMPL model~\cite{SMPL2015} directly from a single RGB image.
% video-based methods
For video-based pose estimation, additional temporal dynamics modeling is usually required to regularize per-frame pose estimation results for stability. 
Existing solutions utilize temporal encoding networks~\cite{Kanazawa_2019_CVPR} %HMMR,
video discriminators~\cite{Kocabas_2020_CVPR} %VIBE
or spatial-temporal attention~\cite{wan2021encoder} %MAED
for temporal regularization.
% performance-capture
Apart from parametric model, single-view performance capture methods~\cite{MonoPerfCap,LiveCap2019tog,DeepCap_CVPR2020} using specific 3D human template can also achieve space-time coherent motion capture.
% free-form performance-capture
While free-form performance capture methods~\cite{Newcombe2015,innmann2016volume, DoubleFusion,robustfusion, su2021robustfusion} get rid of the specific human template but still suffer from the fragile motion field capture.  

% data-driven pose distribution & our method
Among above methods, SMPLify~\cite{keepitSMPL} proposes to optimize estimated SMPL parameters close to the real distribution derived from a Gaussian Mixture Model (GMM), and SMPL-X~\cite{SMPL-X:2019} with further taking gesture and facial expression into account proposes VPoser to replace the GMM model.
VIBE~\cite{Kocabas_2020_CVPR} utilizes an adversarial learning framework that leverages available 3D motion capture dataset to discriminate whether the regressed motion is real or not.
In contrast, our framework adequately exploit the human motion distribution and utilize an expressive latent motion representation to achieve the higher capture accuracy.

\hspace*{\fill} \\{\bf Human Motion Models} are crucial for imposing prior regularization over motion reconstruction. 
% early work (refer to humor & actor)
Early methods introduce various mathematical models to characterize human motion dynamics, including probability model~\cite{howe1999bayesian, grochow2004style, lehrmann2013non}, functional analysis~\cite{ormoneit2005representing} or topological constraints~\cite{urtasun2007modeling}, yet the prior is mostly defined over pose space with limited applicability to simple cyclic motions like walking and running. 
%
% siggraph reviewer
% The very first paper that combines deep learning and character animation proposes a similar idea in spirit- “Learning motion manifolds with convolutional autoencoders” [Holden et al., 2015] learns a manifold of natural human motion using auto-encoder. 
% Then in a follow-up work “A deep learning framework for character motion synthesis and editing” [Holden et al., 2016] it uses this latent space as a motion prior to solve various ill-posed problem in motion synthesis. I was surprised to see that these papers was not even cited.
Recent work~\cite{holden2015learning} learns a manifold of natural human motion using auto-encoder, and its follow-up work~\cite{holden2016deep} this latent space of auto-encoder as a motion prior to solve various ill-posed problems.
Physics-based prior models including foot contact~\cite{Brubaker2009EstimatingCD, rempe2020contact} and kinematics~\cite{Li2019Estimating3M, PhysCapTOG2020, Li2021HybrIKAH} are also widely adopted to derive optimization constraints for controlling global pose and environmental interactions, but usually requires professional knowledge.
% humor: motion vae
% Note that MAVE is cited here!
Recently, deep generative motion models, including GAN-based~\cite{alldieck2018video, Peng2021AMP} and VAE-based ones~\cite{ling2020character, guo2020action2motion} has been widely adopted for data-driven prior learning. 

Our framework is closely related to ACTOR~\cite{petrovich2021action} but aims to model the holistic prior distribution of all human motions instead of action-conditioned ones. Another related work is HuMoR~\cite{rempe2021humor} which also aims to learn an expressive motion model with conditional VAE, but requires time-consuming iterative optimization during inference. In contrast, our motion model is completely feed-forward and outputs motion clips instead of poses, leading to real-time video-based motion reconstruction with superior accuracy.

\section{Overview}
% the meaning of our method, the contribution of our method
Our goal is to learn an expressive latent space that fully captures the prior distribution of human motion at sequence level, and utilize the learned motion prior to facilitate accurate and realistic motion capture from a monocular RGB video. Fig.~\ref{fig:pipeline_all} provides an overview of our full learning framework, which
consists of a transformer-based video encoder that maps the input video to a latent vector $z$ over the prior space, and a motion generator to decode the latent vector to a sequence of motion parameters used for rigging a parametric 3D human mesh model (SMPL~\cite{SMPL2015}). The overview of these two parts are as follows.

%First, a transformed-based variational autoencoder is trained over a large-scale marker-based 3D motion capture dataset~\cite{AMASS_ICCV2019}.
% Our variational motion prior model aims to captures the holistic characteristics of human motion at sequence level and propose several important technical contributions including a novel style-mapping block, a two-stage prior training strategy and loss objective design.

\hspace*{\fill} \\{\bf Prior-guided Motion Generator.}
Our transformer-based VAE learns the temporal human motion characteristics by the supervision of the realistic marker-based 3D motion data \cite{AMASS:ICCV:2019}, which consists of a transformer encoder and decoder.
% \hspace*{\fill} \\{\bf Transformer Encoder.}
This motion transformer encoder maps noisy motion parameters into the same latent space, which leverages temporal motion prior to acquire smooth motions.
Here, we first linearly embed the input motion parameters into a middle space.
With the summation on positional encoding, the transformer block then outputs the predicted mean $\mu$ and std $\sigma$ for the parameterized Gaussian distribution.
Using a reparameterization~\cite{kingma2014auto}, we can sample a latent vector from this distribution. 
And then the decoder part
% \hspace*{\fill} \\{\bf Transformer Decoder.}
called Variation Motion Prior (VMP) generates reliable motion sequences.
Specifically, to strengthen the model capability, we adapt the decoder to a style-based architecture~\cite{karras2019style}, where we first employ a non-linear network mapping before the transformer block.
Moreover, we further use AdaIn as a style injection approach used in style-baed GAN~\cite{karras2019style}, to modulate the statistics of features after normalization in transformer blocks.
%
%We will introduce the training technical details in the next section.

\hspace*{\fill} \\{\bf Transformer-based Video Encoder.}
To explore the temporal information, we propose an effective transformer-based unified motion reconstruction framework that utilizes the temporal features at the sequence level and the spatial information at the image domain. 
%In this framework, we also introduce a novel temporal motion prior, which can benefit the whole framework to capture feasible motion and synthesize new motions from the current prediction.
% backbone
Specifically, with the input video clip,
% image -> feature
we utilize a high-resolution network backbone, HRNet~\cite{sun2019deep} to extract multi-resolution features for each frame. 
To maintain the spatial information on the image domain, we utilize the spatial-temporal transformer, called STE block~\cite{wan2021encoder}, to encode the multi-scale image feature to motion feature for each frame.
% features => mu sigma
We further map the sequential motion features to a parameterized Gaussian distribution, which is used to sample latent code as above for the whole motion clip.
%
% We can directly edit this latent code during video inference, like adding a pre-designed latent difference or interpolating between two predicted latent codes for more relevant motion choices.

\section{Technical Details}
% 下午整体过一遍 参考excel 表格 丰富细节和内容 以及原因

Given an input video clip $V$ of length $T$, our goal is to learn an expressive latent code $z$ and a corresponding motion generator $G$ that maps $z$ to a sequence of motion parameters $\mathbf{P}$ including root translation $\mathbf{r}_t$, pose $\boldsymbol{\theta}_t$ and shape $\boldsymbol{\beta}$ of the SMPL model.
% Our goal is to synthesize natural motions from an input video clip and sample from stylized latent space to generate more relevant motions.
%
Fig.~\ref{fig:pipeline_all} provides an overview of our two-stage learning framework. Stage I propose a variational autoencoder to train the motion prior using the marker-based 3D motion dataset, allowing plausible motions to be generated from a latent variable.
Stage II introduces a transformer-based video encoder to extract the corresponding latent variable from the video input. 
To this end, the video encoder first encodes multi-scale features of each frame $I_{t}$ using a high-to-low resolution CNN backbone~\cite{sun2019deep}.
% features => latent
A spatial-temporal transformer is then adopted to encode sequential feature pyramids into a latent variable $z$ containing the motion information of this clip.
% latent => motion
To decode this latent variable, we leverage our pre-trained variational motion prior to regress motion parameters $\mathbf{P}$ of the SMPL model at each frame.
% editing
% After that, the motion prior is used to generate more relevant motions from the editable latent variable. %(see Sec.~\ref{sec:cap}).
% % Specifically, our framework is described in the following.
In the following, we will introduce the variational auto-encoder for our motion prior (Sec. \ref{sec:vae}), the transformer-based video encoding for our end-to-end framework (Sec. \ref{sec:cap}), and the training strategy (Sec. \ref{sec:train}), respectively.

% Summary: input=>latent=>output
% \cx{novely designed? 具体designed里什么}
% new-designed -> novel?
% More specifically, % 上面提了pre-trained variational motion prior，这块应该是上面的一个具体，所以加了这个，逻辑还可？
% to utilize the motion characteristics for both plausible motion capture and stylized motion synthesis, we introduce the variational motion prior, which is the decoder part of a novel transformer-based variational autoencoder (VAE) in Fig.~\ref{fig:net}.
% %
% To train the motion prior, the transformer VAE first encodes the noised motion input $\mathbf{P}^{noise}$ so as to predict mean and standard deviation (std) for the parameterized Gaussian distribution.
% % latent mapping
% For more robust stylized synthesis, we map the sampled latent code $z$ to $\mathcal{W}$ space with the none-linear mapping block.
% % AdaIN
% Moreover, to strengthen the model capability, we also introduce AdaIN in the transformer to inject latent while the decoding.
% %
% Under the supervision on 3D marker-based data, our variational motion prior thus generates more plausible motion clip (see Sec. \ref{sec:vae}).

\subsection{Stage I: Variational Motion Prior Model}
\label{sec:vae}
\begin{figure*}[t]
    \centering
    \includegraphics[width=\linewidth]{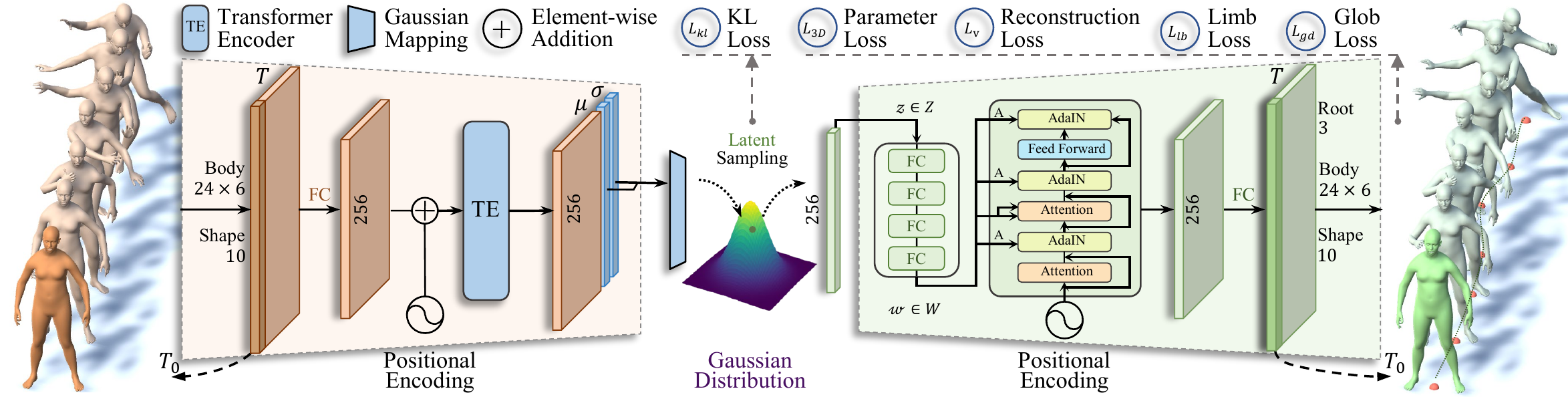}
    % \vspace{5pt}
    \caption{Illustration of our transformer-based variational autoencoder (VAE). Our transformer VAE encodes the temporal motion information to motion latent and decodes this latent to realistic 3D motion clip. Variational Motion Prior (VMP) is introduced with the pre-trained decoder that performs non-linear mapping and AdaIN operation to generate motion sequences from the motion prior space.}
    \label{fig:net}
\end{figure*}

Different from other video-based approaches~\cite{HMR18, Kanazawa_2019_CVPR, kolotouros2019learning, Kocabas_2020_CVPR, wan2021encoder} that directly learn to regress 3D motion parameters from videos, our VMP model aims to capture rich motion prior knowledge from realistic marker-based 3D motion data \cite{AMASS:ICCV:2019} without video supervision. To this end, we propose a transformer-based variational autoencoder to learn the VMP latent space, which can help the subsequent motion capture framework in Sec.~\ref{sec:cap} to generate more natural and plausible motions.
% %
% Note that, same to Eq.~\ref{eq:motion}, we still use $\mathbf{P} = \{\mathbf{p}_t|t\in[1,T]\}$ as the motion representation.
%
Specifically, we formulate the motion parameters $\mathbf{P}=\{\mathbf{p}_{t}| t=[1,T]\}$ of the SMPL model at each frame as follows:
\begin{align}
    \label{eq:motion}
    \mathbf{p}_t= \{\mathbf{r}_t\in\mathbb{R}^3, \boldsymbol{\theta}_t\in\mathbb{R}^{24\times6}, \boldsymbol{\beta}\in\mathbb{R}^{10}\},
\end{align}
which includes root translation $\mathbf{r}_t$, the pose and shape parameters $\boldsymbol{\theta}_t, \boldsymbol{\beta}$ of the SMPL model. Note that the pose representation is converted from the axis-angle format to 6D rotation matrix in ~\cite{zhou2019continuity} for efficient and robust regression.
% todo: 写下 encoder decoder的公式表达和映射关系
As illustrated in Fig.~\ref{fig:net}, our VAE is composed of two modules: motion encoder $\mathscr{M}_{enc}$ and the corresponding motion generator $\mathscr{M}_{dec}$, where detailed module architecture is further described below.
% In the following, we first introduce transformer encoder and transformer decoder of our variational motion prior one by one.

% encoder
% motion clips => nn.linear => motion features
% motion features + \mu + \sigma => transformer decoder => xxx
% sampling z: refer to actor

\hspace*{\fill} \\{\bf Motion Encoder.}
The motion encoder $\mathscr{M}_{enc}$ is designed to generate latent vector $z$ from the input motion parameters $\mathbf{P}$.
We follow ACTOR\cite{petrovich2021action} to build this motion encoder, but aim to capture the holistic motion prior distribution rather than the conditional motion prior based on discrete action categories.
Specifically, we first linearly embed the input motion parameters into a middle space, and this
embedding dimensionality is 256.
We concatenate the learnable distribution parameters $\mu_0 \in \mathbb{R}^{256}$ and $\sigma_0 \in \mathbb{R}^{256}$ with this embedded motion feature.
With the summation on positional encoding, the transformer block then outputs the predicted mean $\mu$ and std $\sigma$ for the parameterized Gaussian distribution.
Using a reparameterization~\cite{kingma2014auto}, we follow the sample of a latent vector $z$ from this distribution. 
Moreover, we apply random noisy for the input motion parameters $\mathbf{P}^{noise}=\{\mathbf{p}^{noise}_t, t\in[1,T]\}$ to enhance the robustness of this VAE for smooth motions.

% decoder
% z => mapping to => w (refer to styleGAN)
% w + position decoder => motion clips
% Refer to:
% Style injection. We first strengthen the model capability by adapting the generator to a style-based architecture [31, 321 as shown in Figure 2(b). We learn a non-linear mapping f: z > W to map the latent code 2 from 2 space to W space, which specifies the styles that are injected into the main synthesis network. We investigate the following style
%
% The encoder learns to predict two vectors, the mean and standard deviation of a distribution. These are then used to parameterize the distribution and generate a sample, z. z is then decoded using the learned decoder.
\hspace*{\fill} \\{\bf Motion Generator.}
Given this latent vector $z$ sampled from a parameterized Gaussian distribution, this motion decoder, as a result, which we called Variation Motion Prior (VMP), generates reliable sequential motion outputs.
Specifically, to strengthen the model capability, we adapt the decoder to a latent mapping architecture~\cite{karras2019style} as shown in Fig. \ref{fig:net}, in which the decoder part lies in the right half of Fig.~\ref{fig:net}.
We first employ a non-linear network mapping from $\mathcal{Z}$ space to $\mathcal{W}$ space before the transformer block.
Comparing to sampling directly from $\mathcal{Z}$ space, the intermediate $\mathcal{W}$ space does not have to support sampling according to fixed distribution, which will be beneficial for the robustness of generated motion.
Moreover, we further use AdaIn as a style injection approach used in StyleGAN generator~\cite{karras2019style}, to modulate the statistics of features after normalization in transformer blocks.
We finally map the middle space features to the predicted motion parameters $\hat{\mathbf{P}}=\{\hat{\mathbf{p}}_t, t\in[1,T]\}$  with a linear projection.
With the direct supervision of motion parameters and indirect supervision on differentiable SMPL layer, VMP thus generates plausible motion clips close to marker-based realistic 3D motion.

\subsection{Stage II: Video Encoder}\label{sec:cap}
Based on the pretrained VMP model that defines an expressive latent space for human motions, we implement our video-based motion capture framework by introducing a video encoder that maps the input video clip to the latent space. 
Here, we propose an effective transformer-based video encoder that utilizes both temporal features at the sequence level and spatial features at the frame level.
Specifically, with the input video clip $V=\{I_{t}| t\in[1,T]\}$,
% image -> feature
We utilize a high-resolution network backbone, HRNet~\cite{sun2019deep} to extract multi-resolution feature maps for each frame $I_{t}$, including $\{1024, 512, 256\}$ dimensions of high-to-low resolution.
To maintain the spatial information on the image domain, we utilize the spatial-temporal transformer, called STE block~\cite{wan2021encoder}, to encode the multi-scale video features to a $256$-dimensional latent motion embedding.
%
% features => mu sigma
We further map the sequential motion features to mean $\mu$ and standard deviation $\sigma$ of a parameterized Gaussian distribution, which is used to sample latent code $z$ and generate the motion parameters $\mathbf{P}=\mathscr{M}_{dec}(z)$ using the pretrained motion generator. 
% 这句表述有些冗余
%For the decoding of this latent code $z$, we present our pre-trained variational motion prior (details in Sec.\ref{sec:vae}) to regress the motion parameters  of the SMPL model at each frame.
% 6D Rotation Representation For Unconstrained Head Pose Estimation
% to modify here
% local 怎么达意？
Note that benefiting from the variational motion prior trained with partial data with global translation, we can directly generate global translation $\mathbf{r}_t$ from local body motion for the complete human movement that \cite{Kocabas_2020_CVPR, wan2021encoder} can not predict accurately.
As for the training of this end-to-end motion capture framework, please refer to Sec. \ref{sec:train} for details.

The latent code in this framework is edible like other image generation work ~\cite{karras2019style}. 
We can directly edit this latent code during video inference, like adding a pre-designed latent difference or interpolating between two predicted latent codes for more relevant motion choices.
The editable latent from the motion prior shows more possibility for our motion capture framework.
We will provide the network design and training details of variational motion prior in the following subsection.

\begin{figure*}[t]
    \centering
    \includegraphics[width=\linewidth]{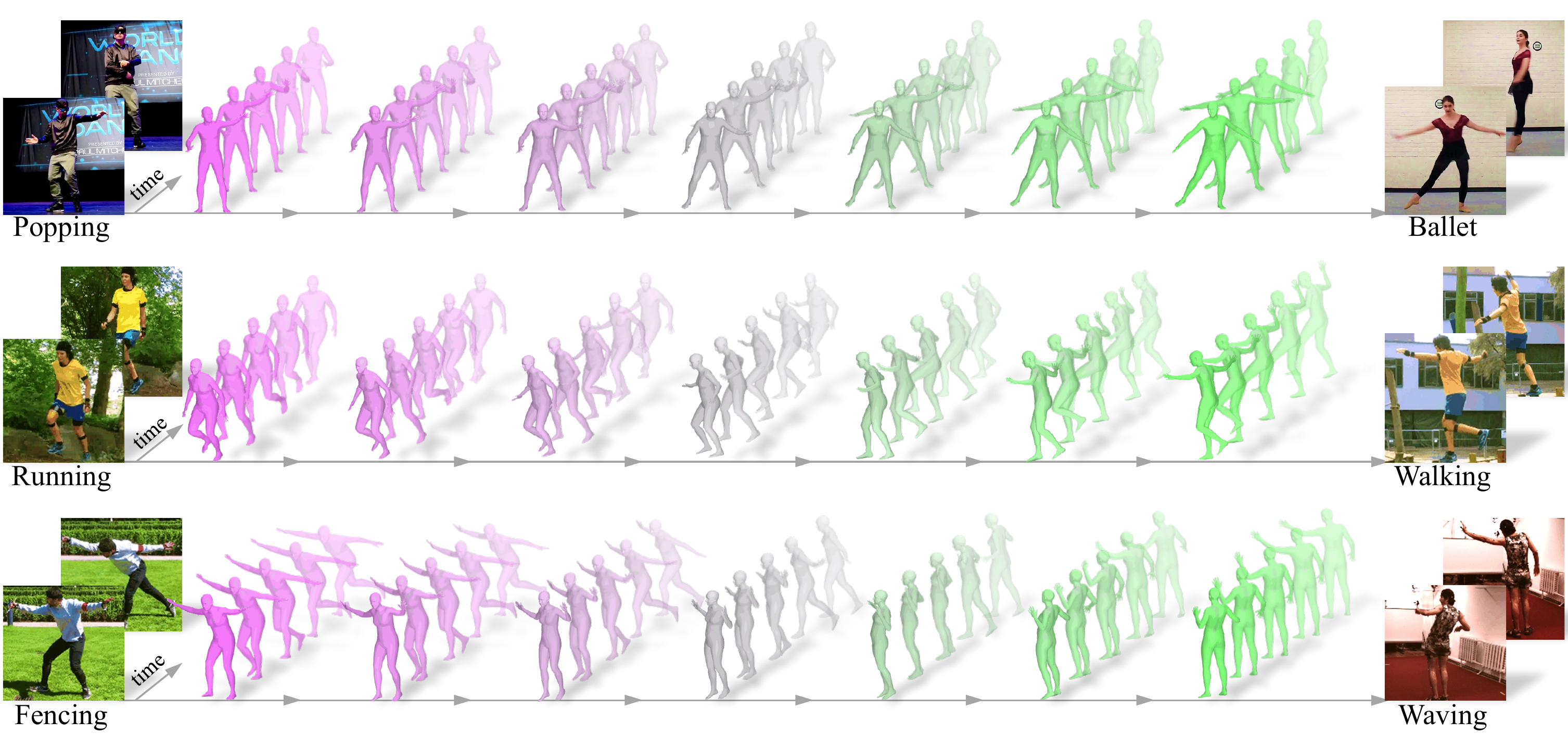}
    \caption{\lingbo{Remove "Style Interpolation" from Fig. 4} Latent space exploration over the pretrained VMP space. The start and end points are defined by encoding two selected video clips with difference motion (the most left and right), and each intermediate motion sequences generated from interpolated latent codes are displayed in colors. Note the smooth transition of motion along the interpolation direction as well as the motion plausibility at each interpolation step.}
    \label{fig_sythesis}
\end{figure*}

\subsection{Training Strategy}~\label{sec:train}

In this subsection, we introduce our training loss and training procedure as follows.

\hspace*{\fill} \\{\bf Loss Function.}
We define the VMP training loss at Stage I as follows:
\begin{align}    
    \label{eq:vae}
    \mathcal{L}_{\mathrm{vmp}} &= \mathcal{L}_{\mathrm{3D}} + \lambda_{\mathrm{lb}}\mathcal{L}_{\mathrm{lb}} + \lambda_{\mathrm{V}}\mathcal{L}_{\mathrm{V}} +  \lambda_{\mathrm{kl}}\mathcal{L}_{\mathrm{kl}}.
\end{align}
% 总体解释，每个loss的目的和作用
where $\mathcal{L}_{\mathrm{3D}}$ encourages the final motion sequence close to the output of network 
% final motion sequence是啥？output又是啥？
% on occluded and invisible joints 
while  $\mathcal{L}_{\mathrm{2D}}$ adds a re-projection constraint on high-confidence 2D keypoints detected from the input images. 
%$\boldsymbol{E}_{\mathrm{T}}$ enforces the final motion to be temporally smooth, while the $\boldsymbol{E}_{\mathrm{S}}$ enforces alignment of the projected 3D model boundary with the detected silhouette.
% We list all losses details in below
Specifically, the 3D \textbf{parameter loss} with \textbf{global} translation is formulated as:
\begin{align}
    \mathcal{L}_{3D} = w_r\sum_{t=1}^T \left\|\mathbf{r}_t  -\hat{\mathbf{r}_t} \right\|_2 + \sum_{t=1}^T \left\|\boldsymbol{\theta}_t  -\hat{\boldsymbol{\theta}_t} \right\|_2 +  \left\|\boldsymbol{\beta}  -\hat{\boldsymbol{\beta}} \right\|_2,
\end{align}
where the three terms represent the global translation $\mathbf{r} \in \mathbb{R}^3$ of the root joint, the pose parameter $\boldsymbol{\theta}_t \in \mathbb{R}^{24\times6}$ and the shape parameter $\boldsymbol{\beta}_t \in \mathbb{R}^{10}$, respectively.
Note that we use $w_r=1$ as default for the training data with root translation but zero for the training data without root translation.
To enhance the supervision on the endpoint of human body, we further formulate the \textbf{limb loss} as:
% 参考sfv
\begin{align}
    \mathcal{L}_{lb} = \sum^T_{t=1}\left\|J_{lb}(\boldsymbol{\theta}_t,\boldsymbol{\beta})-J_{lb}(\hat{\boldsymbol{\theta}_t},\hat{\boldsymbol{\beta}})\right\|_2,
\end{align}
where $J_{lb}$ is the joint regressor that computes 3D joint positions from predicted pose and shape $\boldsymbol{\theta}, \boldsymbol{\beta}$ parameter but extracting only left and right foot/hand joints.
For more specific supervision, we formulate the \textbf{reconstruction loss} on the SMPL vertices:
\begin{align}
    \mathcal{L}_{\mathrm{V}} = \sum_{t=1}^T \left\| \mathbf{V}_t - M(\hat{\mathbf{p}_t}) \right\|_2^2.
\end{align}
Here, the body reconstruction function $M(\cdot)$ is from the differentiable SMPL layer, while the vertices $\mathbf{V}_t$ are calculated with the ground truth motion parameters using the same layer.
The reconstruction loss builds a global supervision on almost all predicted parameters $ \{\mathbf{r}_t, \boldsymbol{\theta}_t, \boldsymbol{\beta}\} $ and shows a reliable supervision~\cite{petrovich2021action} for motion generation. 
For the \textbf{KL loss}, as a standard VAE, we still use Kullback–Leibler (KL) divergence to encourage all generated latent code as a Gaussian distribution.

For Stage-II training of the video encoder, we further introduce the following 2D \textbf{keypoint loss}
% 2D Loss
\begin{align}
    \label{loss:2d}
    \mathcal{L}_{\mathrm{2D}} = \sum^T_{t=1}\sum_{l=1}^{N_J}\phi(l)\left\|\Pi(J(\hat{\boldsymbol{\theta}}_t,\hat{\boldsymbol{\beta}}),s,\mathbf{c})-\mathbf{k}_t^l\right\|_2,
\end{align}  
where the scale and translation parameters of camera is $[s, \mathbf{c}], \mathbf{c} \in \mathbb{R}^2$.
The weak-perspective projection function $\Pi$ maps 3D joint positions to 2D coordinates. 
Besides, the indicator $\phi(l)$ equals to 1 if the confidence $\textbf{C}_{l}$ for the $l$-th joint $\mathbf{k}_t^l$ is larger than 0.5.
The keypoint loss $\mathcal{L}_{\mathrm{2D}}$ encourages the skeleton to match the detection obtained from 2D keypoint estimator, OpenPose~\cite{OpenPose} from the RGB image.  
Finally, combining 2D keypoint loss and VMP prior loss leads to the final learning objective in Stage-II:

\begin{align}
    \label{eq:cap}
    \mathcal{L}_{\mathrm{cap}} &= \mathcal{L}_{\mathrm{vmp}} + \lambda_{\mathrm{2D}}\mathcal{L}_{\mathrm{2D}}
\end{align}

\hspace*{\fill} \\{\bf Training Procedure.}
%For the specific training produce,
As for training procedure, 
we first train the variational motion prior model (Sec.~\ref{sec:vae}) with Eq.~\ref{eq:vae} using Amass~\cite{AMASS:ICCV:2019} dataset and add random Gaussian noise (2 std and 0 mean) on the input motion clips, which takes 12 hours for 200 epochs on a single Tesla V100 GPU.
Freezing this variational motion prior (the decoder parts of VAE), we then train the end-to-end framework (Sec.~\ref{sec:cap}) with Eq.~\ref{eq:cap} using two pose estimation dataset, 3DPW~\cite{vonMarcard2018} and Human3.6M~\cite{h36m_pami}, which takes 2 days for 50 epochs. 
For all following experiments, we use the empirically determined parameters: $\lambda_{kl} = 10^{-5}$, $\lambda_{\theta} = 1$, $\lambda_{lb} = 100$, $\lambda_{2D} = 100$.
We only fine-tune the motion prior part using $10^{-5}$ as the learning rate for the last 5 epochs in Eq.~\ref{eq:cap}, while using $10^{-4}$ for all other training.
The inference speed of the overall framework for capturing motion from the input video is 30 FPS at a single Tesla V100 GPU.

\subsection{Understanding of VMP in Video-based MoCap}
% \cx{to-do 要不要再细化一些？}
% 1. generating more natural motion from monocular video
% 2. sample  between different motion types (in latent space) arbitrarily to generate unconstrained motions
%Based on the pretr variational motion prior, we can utilize both image encoder or motion encoder to regress more natural motion from different model references, like the input monocular videos and the input noisy motion.
The pretrained VMP supports video-based motion capture by projecting input video to the latent space for prior-regularized motion reconstruction. In this section, we provide an in-depth study on the desired properties of VMP, and show how these properties can benefit the video-based motion capture task.

\begin{figure}[t]
	\centering
	\includegraphics[width=\linewidth]{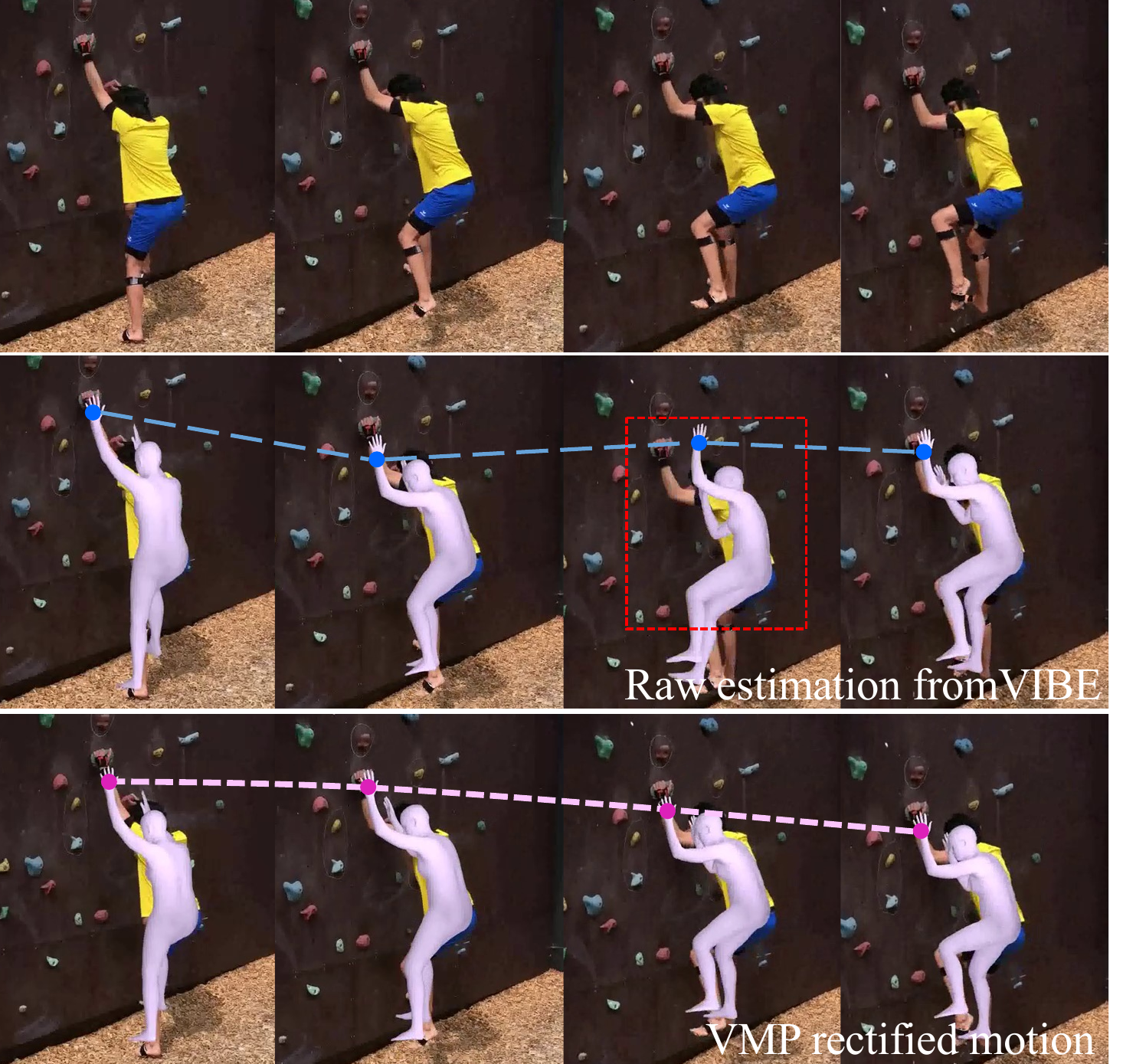}
	% we show more motion capture results from challenging video clips.
	\caption{The pretrained VMP model serve as a motion rectifier that helps generate plausible and temporally stable motion clips.}
	\label{fig_vmp}
\end{figure}

\hspace*{\fill} \\{\bf VMP as a Motion Rectifier.}
A primary focus of motion prior modeling is to promote temporal stability and physical plausibility of output motion clips. Yet existing motion prior regularization are usually affected by the unstable framewise pose estimation. In contrast, our pretrained VMP model essentially provides a motion rectifier that helps correct any pose estimation failures. Here we provide an example of a challenging in-the-wild climbing video where the existing method VIBE~\cite{Kocabas_2020_CVPR} fails, and use the pretrained VAE to reconstruct the correct motion using VIBE's estimation as the input. Fig.~\ref{fig_vmp} compares the noisy motion estimated from VIBE and our rectified results, where the bad estimation in frame 3 is successfully restored after reconstruction, leading to more accurate and temporally stable results. Detailed ablation study results are reported in Sec.~\ref{sec:ablation} to further justify the efficacy of VMP.

\hspace*{\fill} \\{\bf Latent Space Structure Visualization.}
Since our framework is built upon the pretrained motion generator, it is essential that the latent motion space is expressive enough to accurately reconstruct arbitrary motion, and can well represent intermediate motions beyond video-based observations. To verify this, we conduct a latent space interpolation test over the pretrained VMP space, where the start and end points are determined by encoding two randomly chosen video clips. Fig.~\ref{fig_sythesis} visualizes the interpolation results. As can be observed, interpolating between points on VMP space leads to a smooth transition of motion contents, and each intermediate motion is plausible. This indicates that our proposed method successfully learns the latent distribution of real-world human motions.
%we can randomly sample on the latent space to generate arbitrary realistic motions or interpolate a specific latent code different between input motion clip in the latent space of this motion prior.

\section{Experiments}
In this section, we evaluate the efficacy of the proposed variational motion prior at video-based motion capture task. First, we introduce the dataset used for evaluation in Sec.~\ref{dataset}, and validate the superiority of VMP against existing motion prior models in terms of diversity and stability in Sec.~\ref{sec:VMP_evaluation}. Sec.~\ref{sec:VMP_evaluation} reports the motion capture performance against state-of-the-art methods, and the corresponding ablation study results are presented in Sec.~\ref{sec:ablation}. Finally, we perform a user study in Sec.~\ref{sec:user_study} to verify the subjective quality of our method, and discuss the related topics in Sec.~\ref{sec:discussion}.

\begin{figure}[t]
	\centering
	\includegraphics[width=\linewidth]{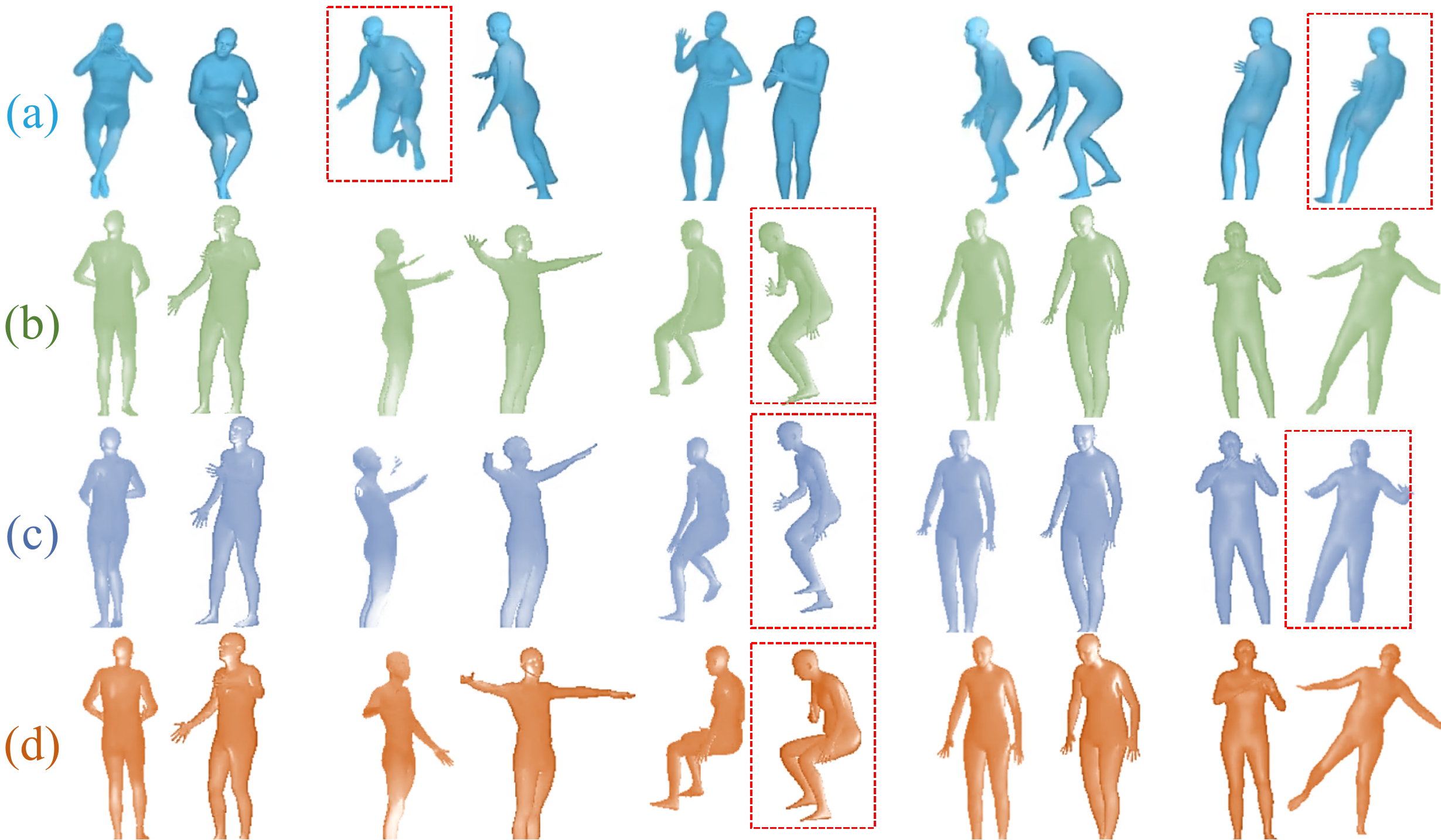}
	\caption{Qualitative evaluation on the diversity and stability on synthesized motion from our VMP. (a)-(d) are synthesized motion samples from ACTOR, ours without non-linear mapping, ours without AdaIn and our full setting, respectively.}
% 	~\cite{petrovich2021action}
	\label{fig_eval_synthesis}
\end{figure}

\begin{table}[t]
	\begin{center}
% 		\vspace{-10pt}
		\centering
		\caption{Quantitative comparison of synthesis diversity. Here, we adpot Average Pairwise Distance (APD)~\cite{aliakbarian2020stochastic} and clip-APD as the diversity indicator for frame-level and clip-level average. To evaluate the diversity from different sampling density, we propose local-APD using std $\sigma$=1 and 5 for Gaussian distributions.}
% 		\vspace{-10pt}
		\label{table:Comparison_sythesis}
		\resizebox{\linewidth}{!}{
			\begin{tabular}{l|ccccc}
				\hline
				\multirow{2}*{Method} & \multirow{2}*{APD$\uparrow$} &  \multirow{2}*{clip-APD$\uparrow$} & local-APD & local-APD\\
				&  & & $\sigma=1 \uparrow$ & $\sigma=5 \uparrow$\\
				% \multirow{2}*{Method} & \multirow{2}{c}{APD$\uparrow$} &  \multirow{2}{c}{clip-APD$\uparrow$} & \multirow{2}{c}{local-APD$\uparrow$} & \multirowcolumn{2}{c}{local-APD}\\ 
				%  Method & APD$\uparrow$  & clip-APD$\uparrow$ &  local-APD($\sigma=1$) $\uparrow$  & local-APD($\sigma=5$)) $\uparrow$
				% \\
				\hline
				HUMOR     &  100  &  -  &  -  & -    \\ % 100 - -
				ACTOR     &  61.4  &  43.5  &  26.6  & 26.6  \\ % 32 * 16
				\cdashline{1-5}[0.8pt/2pt]
				\textbf{Ours}       &  \textbf{122.6}  &  \textbf{75.2 } &  \textbf{27.6} & \textbf{27.0}   \\ % 64 * 80
				\hline
			\end{tabular}
		}
		\vspace{-10pt}
	\end{center}
\end{table}

\subsection{Dataset}\label{dataset}
% overall:
For the dataset, we first use AMASS~\cite{AMASS:ICCV:2019} for training the variational motion prior. We then utilize Human3.6M~\cite{h36m_pami} and 3DPW~\cite{vonMarcard2018} for training and quantitative evaluation of our end-to-end video-based motion capture framework. Moreover, we also introduce the in-the-wild data for further qualitative evaluation.
% one by one:

\noindent\textbf{AMASS}~\cite{AMASS:ICCV:2019} is a large motion capture database containing diverse motions and body shapes on the SMPL body model. 
% use
We sub-sample the dataset to 25 FPS and add random Gaussian noise with 2 std and 0 mean to these 3D motion data as training input of our variational autoencoder. 
%  in Sec. \ref{sec:vae}

\noindent\textbf{Human3.6M}~\cite{h36m_pami} is a large 3D human pose dataset acquired by recording the performance of 5 female and 6 male subjects, under 4 different viewpoints in a fixed indoor environment.

\noindent\textbf{3DPW}~\cite{vonMarcard2018} is an in-the-wild 3D human pose dataset captured based on video and IMU sensors outdoors.
It is widely used in the 3D human pose estimation task thus we quantitatively compare our method against state-of-the-arts on this dataset.

\noindent\textbf{In-the-wild data}
In order to demonstrate our generalization and better compare with others, we also evaluate our approach with the in-the-wild data. Here, we download 67 video clips (2 \textasciitilde 5s with 50fps) from YouTube, including dance, aerobics and figure skating.

\subsection{Evaluation of Motion Prior Space}\label{sec:VMP_evaluation}
% to do: diversity metric
\hspace*{\fill} \\{\bf Diversity.}
A good evaluation perspective for our motion prior is from the aspect of motion synthesis using the latent space. That is to say, the quality and diversity are what we concern.  
We first compare the diversity of our framework with HuMoR~\cite{rempe2021humor} and ACTOR~\cite{petrovich2021action} pretrained on the same AMASS dataset. 
%, which also uses a sequence-level VAE latent space in conjunction with the Transformer design.
We use Average Pairwise Distance (\textbf{APD})~\cite{aliakbarian2020stochastic} as the diversity indicator,
which measures the average joint distance between randomly synthesized motion clips.
We follow HuMoR to randomly sample 50 5s motion clips for each method and compute both frame-level and clip-level average, denoted as \textbf{APD} and \textbf{clip-APD}, respectively.
% vd ariations
% Since APD indicates diversity between different clips, we introduce a diversity indicator within each clip, the mean joint distance between all frames of each clip in 50 5s motions, denoted as \textbf{clip-APD}.
In addition, we propose a new variant of APD to measure local motion diversity by aligning root coordinates within each clip, denoted as \textbf{local-APD}. 
Here we adjust sampling from a Normal Distribution by setting mean $\mu$=0 and std $\sigma$=1 and 5.
%
% In addition, we propose a new variant of APD to measure local motion diversity, denoted as \textbf{local-APD}, by sampling \sz{xxx  % sigma前加上变量表述
% density $\sigma$=1 and 5} and aligning root coordinates within each motion clip.
% % sigma前加上变量表述
% setting  \sz{$\sigma$=1 and 5}.
%Note that we include the global translation of root joints for calculating APD and clip-APD while aligning the root coordinates for all pairs during calculating lim-APD.
%
% For a fair comparison, we retrain ACTOR on AMASS dataset and remove its class condition.
%
As shown in Tab.~\ref{table:Comparison_sythesis}, our approach achieves higher synthesis diversity.
%fig 6
As for comparison on synthesis quality, Fig.~\ref{fig_eval_synthesis} (a) and (d) demonstrate the synthesized motion samples by ACTOR and our method, respectively, in which we can generate more natural and stable results. 
Please kindly refer to the supplemental video for more intuitive examples.
%significantly better synthesis result than ACTOR.

\hspace*{\fill} \\{\bf Stability.}
We evaluate our novel variational motion prior by removing the pre-trained decoder part.
The first row and fifth row of Tab.~\ref{table:eval} demonstrate that with the aid of our variational motion prior, the motion results recovering from the video are more accurate.
Note that the top half of Tab.~\ref{table:eval} is evaluated on 3DPW for verifying our capture performance, and the bottom half is evaluated on AMASS for indicating our capability to handle the motion input.
Beyond this, only with our pre-trained variational motion prior that maps motion to a latent space can the pipeline synthesize novel and diverse movements.
% fig.6: to modify fig.6 from pose estimation to motion synthesis
As shown in (d) of Fig.~\ref{fig_eval_synthesis}, this motion prior can help to generate more visually pleasing results.
As for more qualitative evaluations of our VMP for motion capture, please kindly refer to the supplemental video.

% SIG R5: Thanks for the constructive comment. We will add these suggested metrics. ACTOR uses action-motion datasets like HumanAct12. Thus, we will test our approach using these datasets and evaluate our VAE. These comparisons and their details will be added to the revised paper. Moreover, without annotated action categories in the AMASS dataset, we thus train and test both ACTOR and our approach under same category for the fair comparison on APD. These comparisons and their details will be added to the revised paper.
% SIG R4: - **Motion Editing.** We mainly evaluate diversity with ACTOR to highlight the efficacy in our motion prior learning scheme. As for HuMoR, we failed to reproduce HuMoR's APD evaluation protocol and have to resort to the incomplete APD scores in the original paper. We'll complete Table 2 with human ratings included.

% to do: 
% 1. use HumanAct12 dataset to evaluate diversity?
% 2. complete APD scores in table 2.
% 

\subsection{Comparison}
% begin
In this subsection, we demonstrate the overall performance of our proposed approach by comparing it against other state-of-the-art video-based pose estimation methods, both qualitatively and quantitatively.
% 像，好，多样性, 分别对应下面每个表格
% which indicates the diversity in synthesis, similarity in motion capture and quality in both synthesis and motion capture.

\begin{table}[t]
	\centering
	\caption{Quantitative comparison with SOTA methods on 3D pose estimation. Note we report baseline scores from their original papers on 3DPW~\cite{vonMarcard2018} and Human3.6m~\cite{h36m_pami} dataset.}  	\label{table:Comparison_pose}
	\resizebox{\linewidth}{!}{
	\begin{tabular}{c|cccc|cc}
		\hline
		\multirow{2}*{Method} & \multicolumn{4}{c|}{3DPW} &  \multicolumn{2}{c}{Human3.6m} \\ 
		\cline{2-7}
		& PAMPJPE$\downarrow$  & MPJPE$\downarrow$  & MPVPE$\downarrow$  & ACCEL$\downarrow$ & PAMPJPE$\downarrow$  & MPJPE$\downarrow$\\
		\hline 
	    HMR &  81.3  &  130.0  &  -  &  37.4         & 56.8 & 88.0 \\
	    HMMR &  72.6  &  116.5  &  -  &  -           & 56.9 & -  \\
	    SPIN &  59.2  &  96.9  &  116.4  &  29.8     & 41.1 & -  \\
% 		\cdashline{1-7}[0.8pt/2pt]
		VIBE &  51.9  &  82.9  &  99.1  &  23.4      & 41.5 & 65.6   \\
		MAED &  45.7  &  79.1  &  92.6  &  17.6      & {\bf 38.7} & 56.4 \\
		\cdashline{1-7}[0.8pt/2pt]
		{\bf Ours} & {\bf 44.4} & {\bf 74.2}  & {\bf 82.0} & {\bf 13.4} &  39.1 & {\bf 54.6}\\
% 		{\bf Ours}  & {\bf 45.2} & {\bf 77.0} & {\bf 85.8} & {\bf 11.4} & {\bf 37.5} & {\bf 55.8}  & {\bf ??} & {\bf ???}\\		
		\hline
	\end{tabular}
	}
\end{table}

\begin{figure}[t]
	\centering % 0.95
	\includegraphics[width=\linewidth]{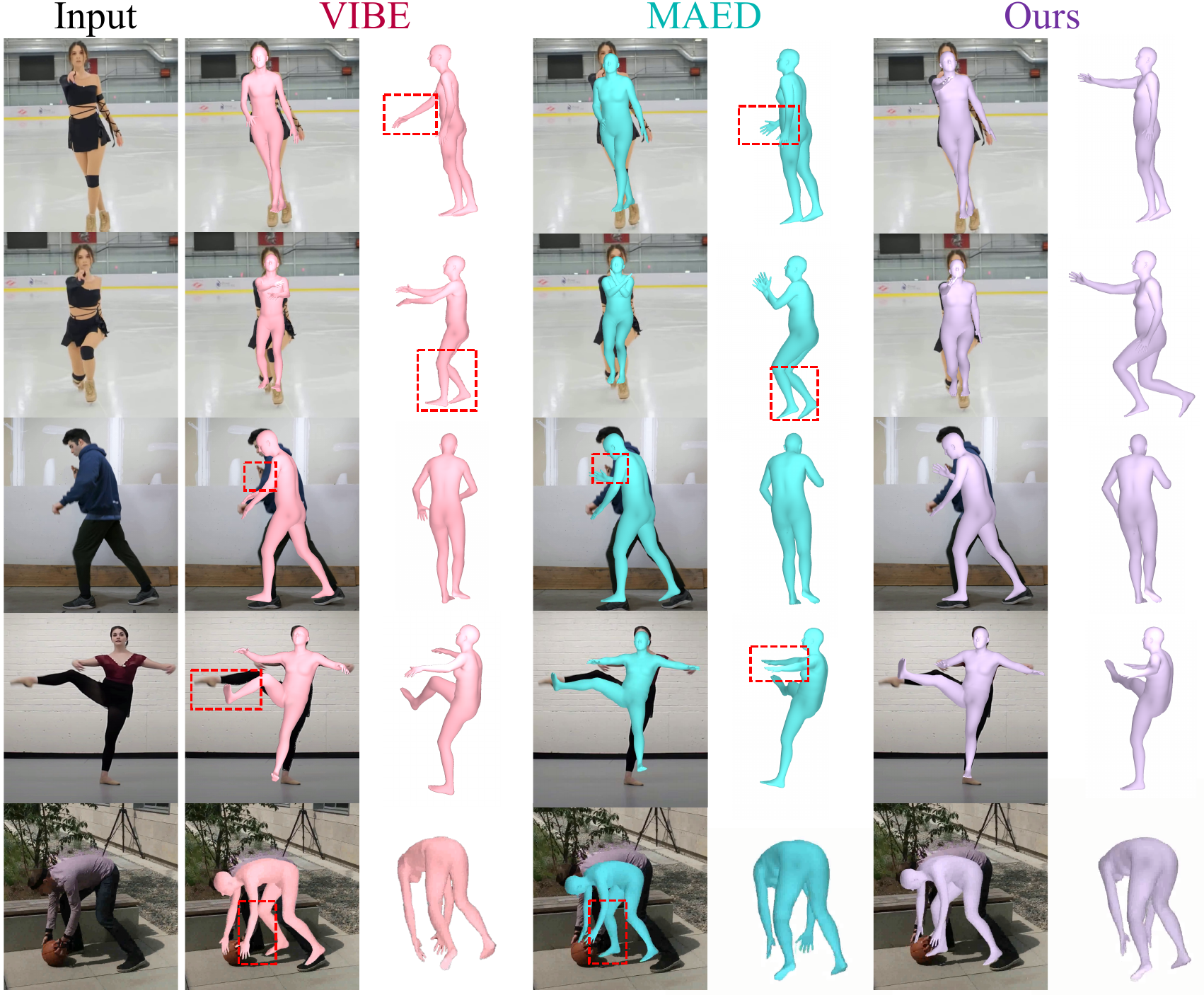}
	\caption{Qualitative comparison on 3D motion capture. Our framework generates more accurate and plausible motions from input video frames against baseline methods.}
	\label{fig_com_pose_estimation}
\end{figure}

\begin{figure}[t]
	\centering 	
	\includegraphics[width=\linewidth]{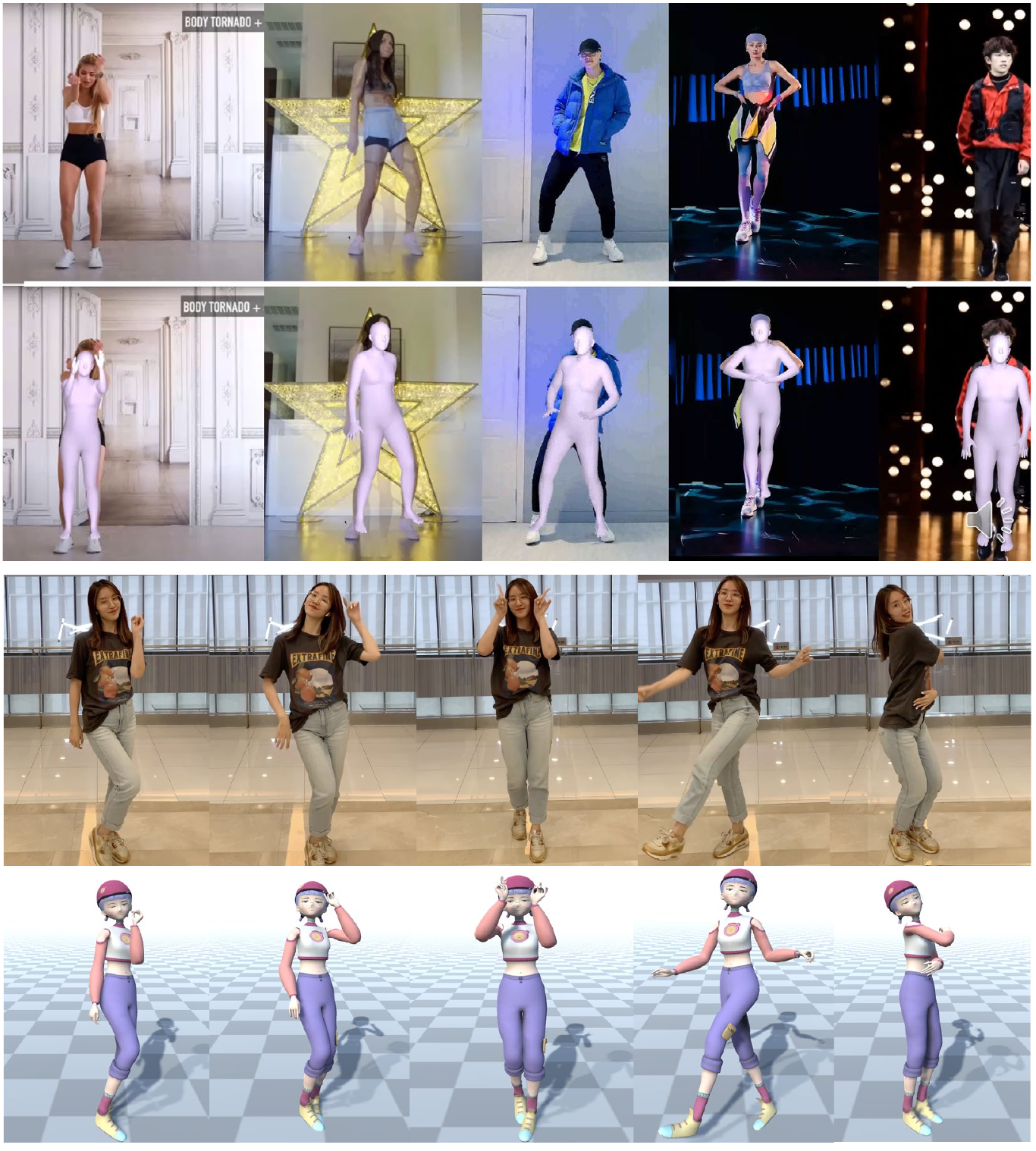}
	% we show more motion capture results from challenging video clips.
	\caption{Top: Randomly selected motion capture results on in-the-wild videos. Bottom: A virtual avatar system designed based on the proposed VMP-based framework for character animation and social media applications.}
	\label{fig_inthewild}
\end{figure}

% \hspace*{\fill} \\{\bf Comparison with pose estimation methods.}
% 
We compare our method with competitive video and image-based estimation methods HMR~\cite{HMR18}, HMMR~\cite{Kanazawa_2019_CVPR}, SPIN~\cite{kolotouros2019learning}, VIBE~\cite{Kocabas_2020_CVPR} and MAED~\cite{wan2021encoder}.
As shown in Fig.~\ref{fig_com_pose_estimation}, video-based motion capture method VIBE~\cite{Kocabas_2020_CVPR} and MAED~\cite{wan2021encoder} suffer from unnatural body lean and self-occlusions, while our VMP-based framework can generate significantly more natural and coherent motions for in-the-wild videos.
%, which is significantly better motion capturing results.
For quantitative evaluation, we report Mean Per Joint Position Error (\textbf{MPJPE}) and its procrustes-aligned variant (\textbf{PA-MPJPE}) on 3DPW~\cite{vonMarcard2018} and Human3.6M~\cite{h36m_pami} dataset. Also, Per Vertex Error (\textbf{PVE}) and ACCELeration error (\textbf{ACCEL}) are listed for comparison.
As illustrated in Tab.~\ref{table:Comparison_pose}, benefiting from the proposed motion prior, our approach achieves the highest accuracy on the challenging 3DPW dataset and is also comparable on Human3.6M with state-of-the-art methods.
Furthermore, as shown in Fig.~\ref{fig_inthewild}, our method also exhibits satisfactory generalization capabilities over in-the-wild videos.

% \subsection{Evaluation}
\subsection{Ablation Study}\label{sec:ablation}
We evaluate our technique components, i.e., variational motion prior, transformer-based network design, HRNet module, non-linear mapping, and training losses.
%
%Moreover, we also evaluate our extension capability by experiment of full-body motion synthesis. 

% SIG R4: 
% - **Global Motion Trajectory.** We'll use other datasets like TotalCapture for quantitative evaluation. Note that both video-based Human3.6M and 3DPW do not provide global data for training. Our ability to predict the global trajectory from video input is transferred from the motion-only AMASS, benefiting from the proposed two-stage prior training.
% - **Extension to Hands.** The jerky wrist issue is mostly caused by motion blur and spatial occlusion under the monocular setting, which will be discussed in the limitation section. More results and quantitative evaluation will also be provided in camera-ready.

\begin{table}[t]
    \Huge
	\begin{center}
		\centering
        % \vspace{-10pt}
		\caption{Quantitative ablation study of technical components our framework on 3DPW (top) and AMASS (bottom). Note that MPJPE calculated on AMASS are with global translation.}
% 		\vspace{-10pt}
		\label{table:eval}
		\resizebox{0.46\textwidth}{!}{
			\begin{tabular}{l|cccc}
				\hline
				Method      &  PAMPJPE  $\downarrow$ 
				&  MPJPE  $\downarrow$ 
				&   MPVPE   $\downarrow$   \\
				% \hline
				% (dataset) & \multicolumn{3}{c}{3DPW} \\
		        \hline 
				w/o motion prior &  55.2   &  87.3  &  102.6        \\								% exps_avatar3d/Regressor/0419_uicap_alldata_600_theta_reverts_kl1_100_trans_regvae
				w/o transformer design &  51.4   &  81.4    &  90.5     \\
				w/o HRNet module &  46.8   &  80.2 &  89.6         \\
			
				w/o KL loss      &  45.1     &  76.2  &  85.9\\
            	\cdashline{1-4}[0.8pt/2pt]
				Ours                  &  \textbf{44.4}   &  \textbf{74.2}  &  \textbf{82.0} \\
				% Ours                  &  \textbf{44.9}   &  \textbf{74.4}  &  \textbf{82.4} \\				
				\hline
				\hline
			 %   (dataset) & \multicolumn{3}{c}{AMASS} \\
		      %  \hline 
				% \hline
				% /apdcephfs/share_1227775/shingxchen/uicap/exps/0428_uicap_alldata_theta_kl_overlap1_16_newlimb_mappingnew/checkpoint_0176.pth.tar
				w/o reconstruction loss      &  37.2     &  50.7 &  58.9\\
				% /apdcephfs/share_1227775/shingxchen/uicap/exps/0428_uicap_alldata_theta_rcverts_kl_overlap1_16/checkpoint_0225.pth.tar
				w/o limb loss        &  33.6   &  45.3 &  54.1 \\
				% /apdcephfs/share_1227775/shingxchen/uicap/exps/0430_uicap_alldataNoise_theta_rcverts_kl_overlap1_16_newlimb_mappingnew_8_fps/checkpoint_0194.pth.tar
				w/o non-linear mapping      &  33.4     &  45.3       &  53.2 \\
				w/o AdaIn      &  34.6     &  46.3      & 53.8 \\
				% \hline
				\cdashline{1-4}[0.8pt/2pt]
				% /apdcephfs/share_1227775/shingxchen/uicap/exps/0430_uicap_alldataNoise_theta_rcverts_kl_overlap1_16_newlimb_mappingnew/checkpoint_0054.pth.tar
				Our VAE      &  \textbf{31.6}   &  \textbf{42.1} &  \textbf{49.4}  \\
				\hline
			\end{tabular}
		}
% 		\vspace{-10pt}
	\end{center}
\end{table}

\hspace*{\fill} \\{\bf Network structure.}
%1. Transformer-based motion capture.
For evaluation of our overall framework, we replace our transformer encoder with a plain ResNet backbone for a method variation without transformer design.
For a fair comparison, we train it using the same strategy as in Sec~\ref{sec:train}.
As shown in Tab.~\ref{table:eval}, we can see that our transformer-based backbone can achieve better capturing accuracy than the one without the transformer design.
%2. HR net
In this table, our introduced HRNet that extracts multi-resolution feature for each frame is also evaluated by removing it in the third row.
%3. others: non-linear mapping AdaIn
For the more detailed part, the non-linear mapping and AdaIn introduced in the VAE also help us to strengthen the model capability, as illustrated in 6th and 7th rows.
Note that the upper half of the Tab.~\ref{table:eval} evaluates the errors of the overall pipeline from video to 3D poses on 3DPW, while the second half
%of the Tab.~\ref{table:eval} 
evaluates the errors for the pre-trained VAE from the motion to synthesized motion on AMASS.
%fig 6
For qualitative evaluation of diversity and stability on motion synthesis, as shown in the (b) and (c) of Fig.~\ref{fig_eval_synthesis}, we also implement the ablation study for the non-linear mapping and AdaIn to demonstrate their effectiveness.

% SIG R2
% **(2) Ablation Studies.** Our AdaIN module shows the most performance improvement, comparing to ACTOR. Meanwhile, the limb loss is quite important for qualitative results. Besides, “w/o motion prior” indicates the training without the pre-trained prior model. “w/o transformer design” indicates replacing our transformer with a ResNet backbone. 
% We'll add ACTOR and the configuration details in the ablation studies for clarification.
% to do: add ACTOR and configuration details here!

\hspace*{\fill} \\{\bf Ablation study for training loss.}
Here, we further investigate the influence of the training loss formulation when using the SMPL model in our network.
We experiment without (i) the particular KL loss $L_{kl}$, (ii) the reconstruction loss $L_{V}$, (iii) the limb loss $L_{lb}$.
In Tab.~\ref{table:eval}, we observe that the training without $L_{V}$ or $L_{lb}$ is not sufficient to constrain the problem and thus leads to inaccurate results, while the combination of the above losses, as our supervision, significantly improves the motion synthesis results.
We thus utilize realistic motions as our supervision.
% 
%Note that KL loss $L_{kl}$ help the motion prior to generate more natural motions while may introduce inaccurate fitting to the 2D images in motion capture problem. Therefor, we eliminate it during the overall end-to-end training process.

\subsection{User Study}
\label{sec:user_study}
% naturality
% diversity
% Beyond the qualitative and quantitative comparison with the motion capture methods, we our VMP model can 
% , we add the user study. Specifically, we ...
%
Beyond the quantitative comparison results above, we also conduct a user study to validate the subjective quality of our VMP against three baselines, namely ACTOR, ours without AdaIN, and ours without non-linear mapping. Specifically, 25 participants are invited to rate the naturalness and realism of the given motion sequences using five-level Liker scale, (i.e., 5: strongly agree; 4: agree; 3: neither agree nor disagree; 2: disagree; 1: strongly disagree), and compute mean opinion score (MOS) for each method. We randomly select 15 synthesized motion samples from latent motion space of each method and shuffle the combined sample list before testing. Our method achieves 3.99 MOS score with 1.5 gain over ACTOR~\cite{petrovich2021action} (2.49 MOS score), thus proving the efficacy of our method. Furthermore,
ours without AdaIN and ours without non-linear mapping get 3.74 and  3.73 MOS score, respectively, which indicates the efficacy of proposed AdaIN block and non-linear mapping. Overall, the efficacy of our motion prior is further justified.
%involving xx male and yy female participants.
% Specifically, we recruit xx people from our company including half female and half male.
% 别透露我们是公司投稿哈
% to do: user study!
% ref: https://citeseerx.ist.psu.edu/viewdoc/download?doi=10.1.1.720.2510&rep=rep1&type=pdf

\subsection{Discussion}\label{sec:discussion}
\begin{figure}[t]
	\centering
	\includegraphics[width=0.9\linewidth]{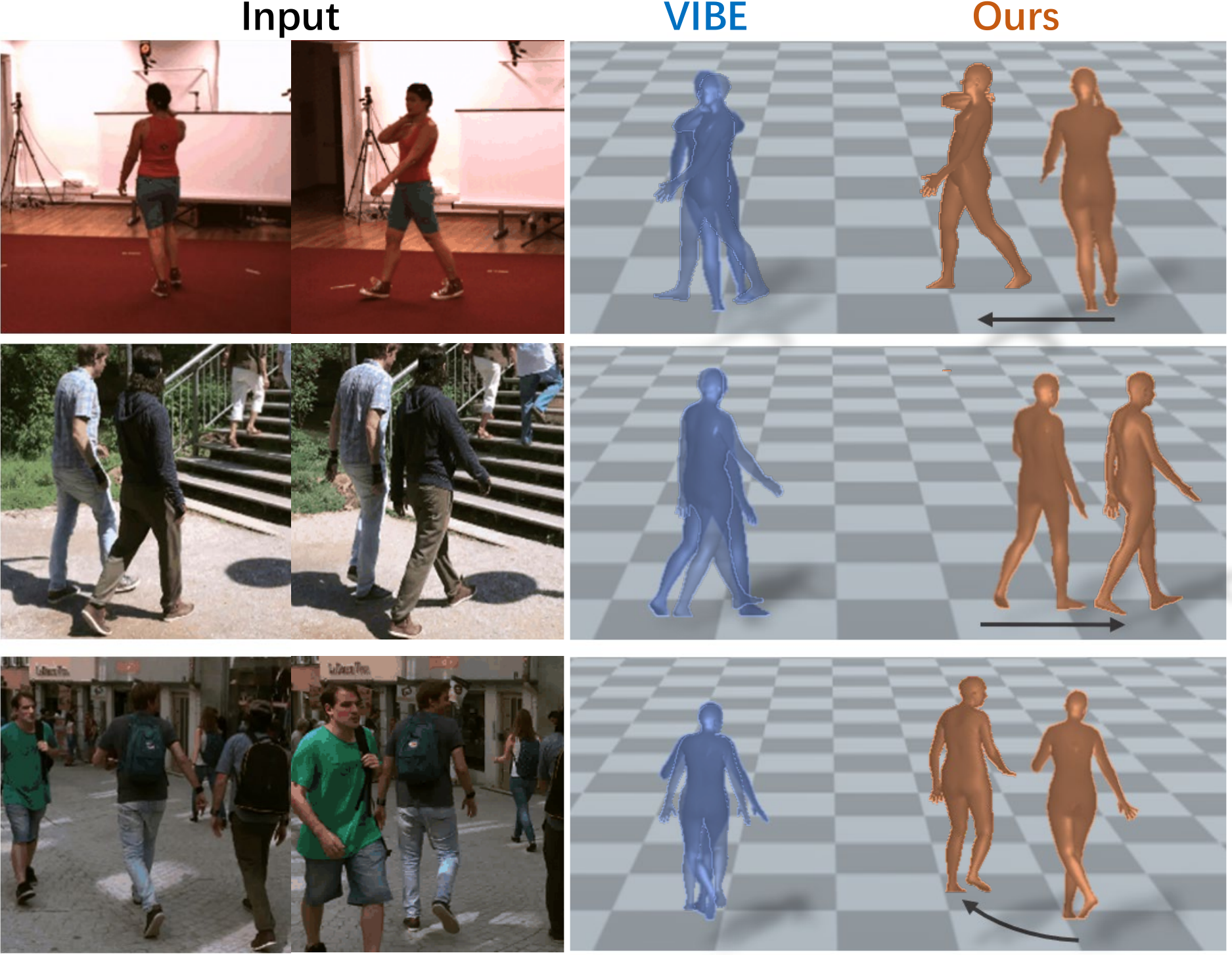}
	\vspace{-5pt}
	\caption{Comparison on the global translation with VIBE. Benefiting from our robust motion prior. We can predict global translation from our end-to-end learning framework without other optimization.}
	\label{fig_global_trajectory}
	\vspace{-10pt}
\end{figure}

Beyond our framework design and the compelling motion capture results demonstrated above, there are still something to be discussed or improved. 
% \hspace*{\fill} \\{\bf Global Motion Trajectory}
First, since both video-based Human3.6M and 3DPW do not provide global trajectory for training, our video encoder cannot encode the global motion trajectory information, like other video-based motion capture methods VIBE~\cite{Kocabas_2020_CVPR} and MAED~\cite{wan2021encoder}.
However, our variational motion prior are trained using the motion-only AMASS with global translation. Benefiting from the proposed two-stage prior training, our framework can generate significantly more natural and coherent motions with a certain degree of global trajectory capture ability, see Fig.~\ref{fig_global_trajectory}.
% discussion
% \hspace*{\fill} \\{\bf Extension to full-body motion synthesis}
Second, although targeted at human body motion, our framework does not rely on any topological constraint and can be easily extended to other articulated objects such as hands.
In future, we expect our method to capture body, hand and face motions in a unified fashion, which can greatly benefit various VR applications.
Besides, our variational motion prior relies heavily on the training data, 
therefore, synthesized motion sequences are more related to motion distribution of the training dataset and a more comprehensive MoCap dataset with more complex motions would greatly boost the performance.
Also, our prior can only handle singe-person, extending our motion prior model to multi-person interactions is another promising direction.
Finally, our variational motion prior is trained purely on local motion features, making it less effective in absolute location prediction, such as distinguishing if a person is walking on a treadmill. Combing physical and motion prior constraints is another possibility.
% Moreover, it is also a promising direction for total human motion synthesis including body, hand and face. 
% Because our variational autoencoder framework is easily expanded to different topology, we can also \cx{recover} the hand motion from the input video based on hand capture method~\cite{PIXIE:2021} and our hand motion prior. 

\section{Conclusion.}

We propose a novel motion prior learning framework to facilitate video-based human motion capture with superior accuracy and stability.
% 主贡献
Our key contribution lies in the variational motion prior (VMP) formulation that learns an expressive latent representation of human motions, 
% 优越性
allowing more plausible motion reconstruction from video input as well as better generalization ability towards challenging in-the-wild videos with complex motions.
% 总模块
We substantiate VMP via a transformer-based VAE trained over 3D MoCap data, allowing rich motion prior to be utilized for the video-based motion capture task.
% 实验总结
Extensive experiments over both benchmark datasets and in-the-wild videos have demonstrated the efficacy and versatility of our framework, 
% 具体好在哪里
with superior accuracy and plausible motion reconstruction against state-of-the-art motion capture methods. 
We believe our approach can serve as a critical step for robust monocular human mocap with the aid of generative motion prior, with numerous potential applications
in VR/AR like gaming and entertainment.
%\hspace*{\fill} \\{\bf Limitations and Future Works}
% \newpage

%% if specified like this the section will be committed in review mode
\acknowledgments{
The authors wish to thank A, B, and C. This work was supported in part by
a grant from XYZ (\# 12345-67890).}

\bibliographystyle{abbrv-doi}

\bibliography{template}

\end{document}